\documentclass[a4paper]{spie}  %>>> use this instead for A4 paper
\usepackage{amsmath,amsfonts,amssymb}
\usepackage{graphicx}
\usepackage{algorithm}
\usepackage{algorithmic}
\usepackage{bm}
\usepackage[colorlinks=true, allcolors=blue]{hyperref}

\newcommand{\figref}[1]{Fig.~\ref{#1}}

\newcommand{\tabref}[1]{Tab.~\ref{#1}}

\newcommand{\equaref}[1]{Eq.~\eqref{#1}}

\title{Interactive image diagnosis using test-time backpropagation from the auxiliary task}

\author{Takuro Shimaya}
\affil{Biometrics Research Laboratories, NEC Corporation, Kawasaki, Japan}

\authorinfo{Further author information: (Send correspondence to T.S.: E-mail: takuro-shimaya@nec.com)}

\begin{document} 
\maketitle

\begin{abstract}
    Multi-task deep learning has been developed to ensure explainability in computer-aided diagnosis (CAD) systems.
    These systems simultaneously predict interpretable characteristics of findings in auxiliary tasks to help radiologists assess the reliability of the diagnostic prediction.
    However, the CAD's opinion would not be referred to when the auxiliary task prediction deviates from the radiologist's expectation.
    To further leverage CAD's knowledge in the above situation, we proposed a multi-task gradient assist (MGA) method that brings interactivity to multi-task models without additional training.
    MGA allows radiologists to give feedback to the CAD and ask for additional diagnostic suggestions.
    Using two datasets of mammography and CT, we show that MGA successfully improves diagnostic accuracy upon feedback on various auxiliary tasks.
    Compared to an existing method using forward propagation only, MGA performs more robustly and can be combined to further improve accuracy.
    Our method paves a promising way for integrating artificial intelligence into image diagnosis.
\end{abstract}


\section{Introduction}
\label{sec:intro}  % \label{} allows reference to this section
Radiologists evaluate the malignancy of findings by taking into account the visual characteristics of lesions, e.g., the shape and the margin.
To put it in machine learning terms, radiologists perform multi-task inference, where the main and the auxiliary tasks are malignancy classification and visual characterization of lesions, respectively.
The idea of multi-task deep learning has indeed been proposed for various applications in medical image computing\cite{Zhao2023,Zhang2020,Zhang2021,Zhang2023,Yao2023}, showing that learning to predict the auxiliary task improves the performance of the main diagnostic task.

Multi-task learning not only enhance the diagnostic accuracy of the deep learning model itself but also brings explainability.
With current regulatory approval for medical software, a diagnostic decision cannot be made solely by the computer-aided diagnosis (CAD) model and must be done by radiologists\cite{Brady2024}.
This process would be facilitated if the CAD simultaneously predicts the interpretable auxiliary task, e.g., the lesion margin category, as well as the malignancy classification. 
For instance, radiologists can accept the diagnostic decision given by the CAD only when the prediction on the auxiliary task is acceptable.
This strategy may reduce automation bias, i.e., the tendency of human users to over-rely on the CAD. 

\begin{figure}[t]
    \begin{center}
    \begin{tabular}{c} %% tabular useful for creating an array of images 
    \includegraphics[width=\linewidth]{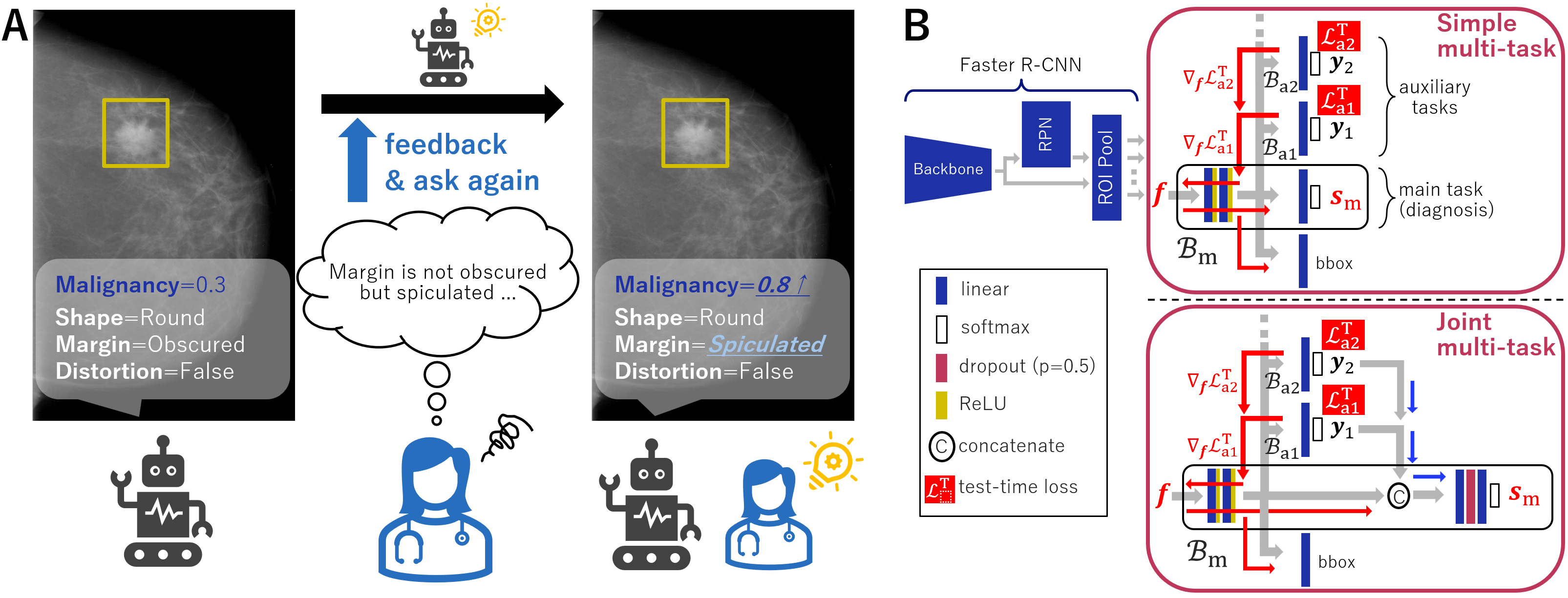}
    \end{tabular}
    \end{center}
    \caption{\label{fig:1} 
    Overview of multi-task gradient assist (MGA).
    (a) An illustration of the applied case.
    The multi-task CAD system receives feedback from the radiologist to refine the prediction.
    (b) The architecture of the networks used in this study.
    The malignancy score $\bm{s}_\mathrm{m}$, the bounding box, and auxiliary tasks $\mathrm{a}_i$ are simultaneously predicted from the ROI feature $\bm{f}$.
    MGA updates $\bm{f}$ via the test-time backpropagation (indicated by the red arrows and see also \equaref{eq:1}).
    The blue arrows indicate the replacement (REPL) of the auxiliary task prediction with feedback, as expressed in Eqs.~\eqref{eq:2} and \eqref{eq:3}.
    }
\end{figure}

On the other hand, this means that typical multi-task CAD systems cannot help radiologists when the auxiliary task prediction deviates from the radiologist's expectation.
For utilizing knowledge from the CAD system even in the above situation, interactivity would be further desired for the CAD system to receive the radiologist's feedback on the auxiliary task prediction and refine the main diagnostic prediction thereby (\figref{fig:1}a).
However, such potential interactivity of multi-task CAD systems remains largely unexplored. 
A multi-task CAD model proposed by Zhang et al.\cite{Zhang2021,Zhang2023} is one of few systems with interactivity.
They employed a joint multi-task architecture (\figref{fig:1}b, bottom), where auxiliary task predictions are directly concatenated with image features and used for malignancy classification.
Although radiologists can directly intervene in the model by replacing the auxiliary task predictions ($\bm{y}$ in \figref{fig:1}b, bottom) to update the main diagnostic prediction, the contribution might be limited because image features, which remain unchanged, dominate the intermediate features used for diagnosis\footnote[1]{Note that Zhang et al. did not explicitly discuss the interactivity of their method.}. 

This study proposes a multi-task gradient assist (MGA) method that implements interactivity to already trained multi-task CAD models without additional training processes.
Our work has three key contributions:
\begin{enumerate}
    \item Our experiments, using two datasets of mammography and lung CT, showed that MGA improves the diagnostic accuracy of the trained multi-task CAD model by teaching the ground truth of interpretable auxiliary tasks such as lesion margin classification. 
    \item Compared to an existing method based on the joint multi-task architecture, MGA performs more robustly and can be combined with the existing method to further improve accuracy upon feedback.
    \item We confirmed a clinically valid behavior of our method: the malignancy score was enhanced when feedback indicated the finding evident in malignant lesions.
\end{enumerate}

\section{Methods}
\label{sec:sections}
\subsection{Multi-task gradient assist (MGA)}
The essence of our method lies in the feedback-driven refinement of the image feature $\bm{f}$, which remains unchanged in the existing framework\cite{Zhang2021,Zhang2023} (blue arrows in \figref{fig:1}b, bottom).
Specifically, we leverage the gradient of the test-time loss $\mathcal{L}^\mathrm{T}_{\mathrm{a}_i}(\bm{y}_i, \tilde{\bm{y}_i})$ on the auxiliary task $\mathrm{a}_i$ with respect to $\bm{f}$, which can be calculated from the error between the prediction by the trained multi-task model $\bm{y}_i$ and radiologist's feedback $\tilde{\bm{y}_i}$.
Note that the gradient $\nabla_f \mathcal{L}^\mathrm{T}_{\mathrm{a}_i}$ is not used for tuning weights of the multi-task model but for changing $\bm{f}$ during the test-time, when radiologists interpret the prediction from the model.
With $\mathcal{B}_\mathrm{m}$ as the sub-network predicting the malignancy score $\bm{s}_\mathrm{m}$ from the image feature $\bm{f}$ (\figref{fig:1}b, top), MGA re-calculates $\bm{s}_\mathrm{m}$ according to radiologist's feedback as follows:
\begin{equation}
    \label{eq:1}
    \bm{s}_\mathrm{m} = \mathcal{B}_\mathrm{m}(\bm{f}) \to \mathcal{B}_\mathrm{m}(\bm{f} - \sum_i \alpha_i \nabla_f \mathcal{L}^\mathrm{T}_{\mathrm{a}_i}(\bm{y}_i, \tilde{\bm{y}_i})),
\end{equation}
where $\alpha_i$ is the hyperparameter to be determined by validation.
Similarly, one can refine other outputs, such as the bounding box regression for the detection model.
Our method is based on adversarial/assistive perturbation\cite{Goodfellow2015,Pestana2021}, which can confuse/assist the trained single-task network by adding the gradient to input images.
MGA aims to selectively enhance the feature relevant to feedback so that $\bm{f}$ gets closer to what the radiologist expects.

Our method is combinable with the existing joint multi-task architecture (\figref{fig:1}b, bottom).
The existing method can directly replace (REPL) $\bm{y}_i$ to re-calculate $\bm{s}_\mathrm{m}$ as
\begin{equation}
    \label{eq:2}
    \bm{s}_\mathrm{m} = \mathcal{B}_\mathrm{m}(\bm{f}, \bm{y}_1=\mathcal{B}_1(\bm{f}), \cdots, \bm{y}_K=\mathcal{B}_K(\bm{f})) \to \mathcal{B}_\mathrm{m}(\bm{f}, \tilde{\bm{y}_1}, \cdots, \tilde{\bm{y}_K}),
\end{equation}
where $\mathcal{B}_i$ is the sub-network for the auxiliary task $\mathrm{a}_i$, and $K$ is the number of total auxiliary tasks.
MGA can be incorporated with REPL and further reflect radiologist's feedback via the test-time backpropagation:
\begin{equation}
    \label{eq:3}
    \bm{s}_\mathrm{m} = \mathcal{B}_\mathrm{m}(\bm{f}, \mathcal{B}_1(\bm{f}), \cdots, \mathcal{B}_K(\bm{f})) \to \mathcal{B}_\mathrm{m}(\bm{f} - \sum_i \alpha_i \nabla_f \mathcal{L}^\mathrm{T}_{\mathrm{a}_i}(\bm{y}_i, \tilde{\bm{y}_i}), \tilde{\bm{y}_1}, \cdots, \tilde{\bm{y}_K}).
\end{equation}

\subsection{Datasets and Experiments}
We utilized two open medical datasets, the CBIS-DDSM\cite{Lee2017} and the LIDC-IDRI\cite{Armato2011}. 
For the CBIS-DDSM, the mammography dataset with ROI annotations, we used images with biopsy-confirmed malignant and benign masses, each of which has four class shape categories, five class margin categories, and the BI-RADS assessment score ranging from one to five.
For the assessment score, we considered BI-RADS=2 and BI-RADS=3 to be in the same category to account the imbalance in the number of cases.
Note that BI-RADS = 1 indicating the absense of suspicious findings was not given to each ROI annotation. 
We also extracted the binary categories indicating the presence of distortion.
In summary, the main task for this dataset is the detection of biopsy-confirmed malignant masses, with four potential auxiliary tasks: the shape, the margin, the assessment, and the distortion classifications (see also \tabref{tab:1a} for the list of categories in the auxiliary tasks).
We finally obtained 983 images for train, 248 for validation, and 361 for test, with 912 benign and 784 malignant masses.

For the LIDC-IDRI, the lung CT dataset, we extracted 2D slices where the center of the nodule is located.
For each nodule with four multiple annotations, the ground truth bounding box and finding categories were obtained by majority voting.
Since biopsy-confirmed diagnosis was absent, we regarded nodules with an average assessment score higher than three as highly suspicious cases.
The main task is the detection of highly suspicious nodules, with five potential auxiliary tasks: the calcification, the margin, the lobulation, the spiculation, and the texture classifications, each of which has five classes.
We finally obtained 1874 images for train, 353 for validation, and 398 for test, with 1351 slightly and 1263 highly suspicious nodules.
\begin{table}[t]
\caption{\label{tab:1} 
Results on the accuracy of malignant lesion detection.
(a) Results on the CBIS-DDSM, mammography (MMG), using the simple multi-task (top) and the joint multi-task architecture (bottom).
Stand-alone represents the detection accuracy of the model without feedback.
MGA, REPL, and MGA+REPL re-calculate the malignancy score of predictions with feedback as Eqs.~\eqref{eq:1}, \eqref{eq:2}, and \eqref{eq:3}, respectively.
Ablation studies were performed on MGA by giving random feedback.
The colored numerals represent the best results for each condition, where blue and red indicate improvement and deterioration from the single-task baseline, respectively.
(b) Results on the LIDC-IDRI, thoracic computed tomography (CT).}
\begin{center}
\begin{tabular}{c} 
\includegraphics[width=\linewidth]{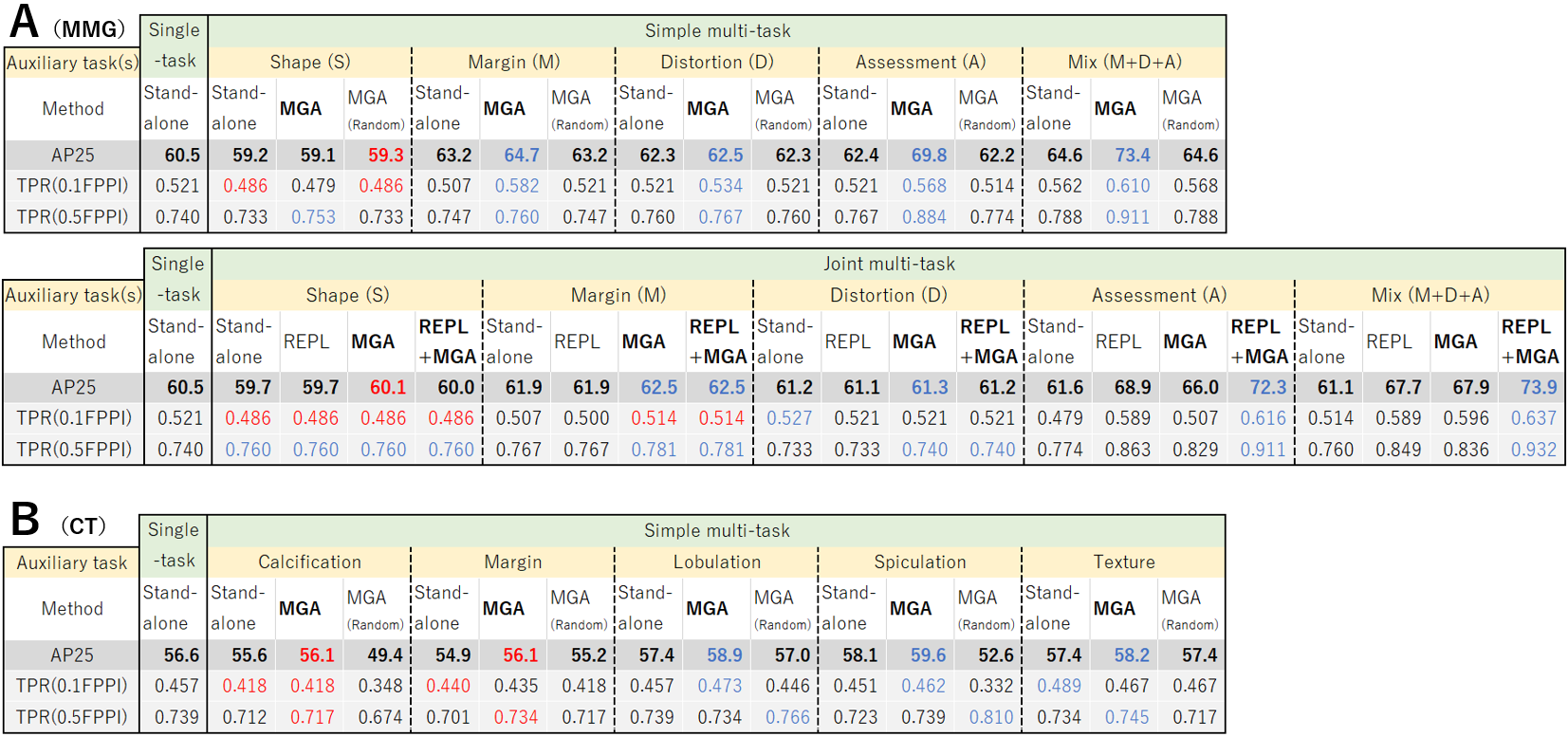}
\end{tabular}
\end{center}
\end{table} 

For the multi-task CAD model, we employed the Faster R-CNN\cite{Ren2015} with the ConvNeXt-T\cite{Liu2022} and FPN\cite{Lin2017} backbone, where the ROI heads have multiple branches to infer auxiliary tasks simultaneously (\figref{fig:1}b).
The model was trained in a similar manner as in ref.\cite{Ren2015}, but the cross entropy loss $\mathcal{L}_{\mathrm{a}_i}$ on the auxiliary task $\mathrm{a}_i$ were added with the hyperparameter coefficient $\lambda_i$.
We defined an independent multi-task model for each auxiliary task, i.e., each model has only one auxiliary task ($K=1$).
In addition, we investigated a model with multiple auxiliary tasks, including margin, distortion, and assessment classifications for the CBIS-DDSM.

After training the multi-task models, we performed MGA and REPL by referring to the ground truth in the validation and test datasets as radiologist's feedback $\tilde{\bm{y}_i}$.
Specifically, we re-calculated the bounding box and the malignancy score $\bm{s}_\mathrm{m}$ of the prediction overlapped with the ground truth malignant and benign ROI with the IoU higher than 0.25.
We also carried out the ablation study on MGA by giving random feedback, where $\tilde{\bm{y}_i}$ was displaced by a random class label.
The detection performance was then compared before and after feedback was given.
We evaluated the accuracy by the average precision with the IoU threshold of 0.25 (AP25), and the true positive rate (TPR) at 0.1 and 0.5 false positives per image (FPPI).
\begin{table}[t]
\caption{\label{tab:1a} 
The confusion matrices on the auxiliary task classification for the CBIS-DDSM.
We evaluated simple multi-task models with only one corresponding auxiliary task, i.e., the four multi-task models with shape, margin, assessment, and distortion classification, respectively (see the upper table in \tabref{tab:1}a).
The numerals represent the number of ground truth masses with corresponding predicted labels.
}
\begin{center}
\begin{tabular}{c} 
\includegraphics[width=\linewidth]{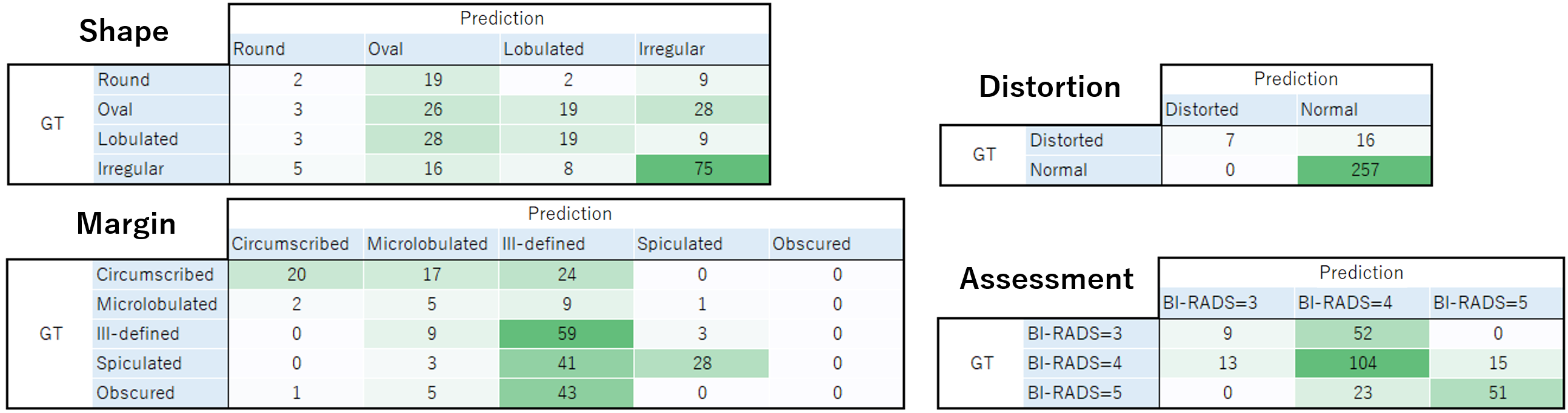}
\end{tabular}
\end{center}
\end{table}

\begin{figure}[t]
    \begin{center}
    \begin{tabular}{c} %% tabular useful for creating an array of images 
    \includegraphics[width=\linewidth]{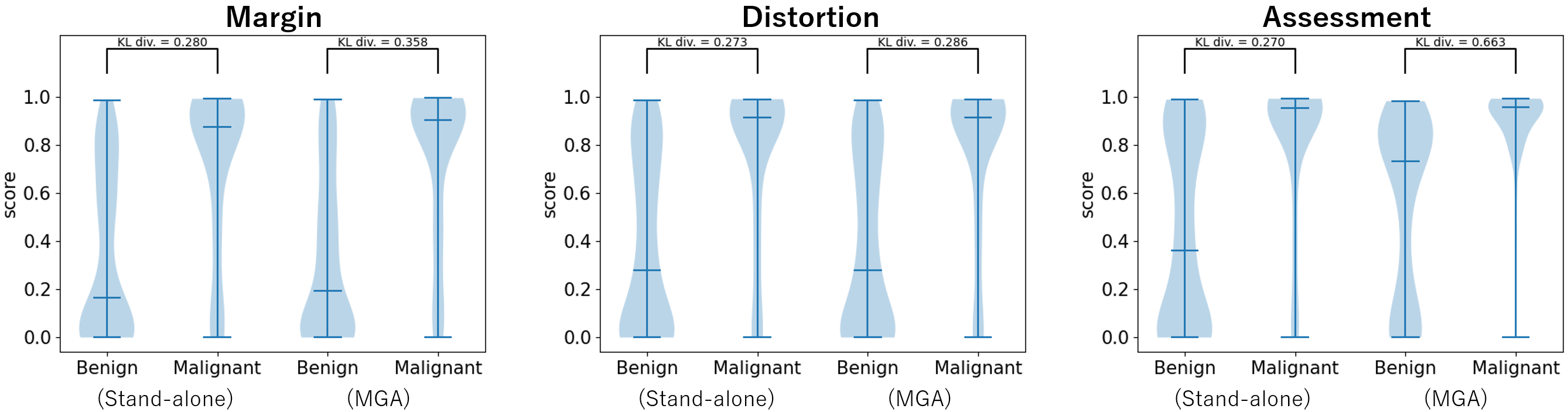}
    \end{tabular}
    \end{center}
    \caption{\label{fig:2} 
    The malignancy score distributions for benign and malignant masses for the CBIS-DDSM.
    We compared the score distributions before and after MGA was applied.
    We evaluated the simple multi-task models with only one corresponding auxiliary task, i.e., the three multi-task models with margin, assessment, and distortion classification, respectively (see the upper table in \tabref{tab:1}a).
    The difference between score distributions of benign and malignant masses, which was quantified by the Kullback-Leibler divergence (KL div.), was enhanced by MGA ($0.280\to 0.358$ for margin, $0.273\to 0.286$ for distortion, and $0.270\to 0.663$ for assessment).
    The KL divergence was calculated as $\int\mathrm{d}\bm{s}_\mathrm{m} P_\mathrm{B}(\bm{s}_\mathrm{m}) \log(P_\mathrm{B}(\bm{s}_\mathrm{m})/P_\mathrm{M}(\bm{s}_\mathrm{m}))$, where $P_\mathrm{B}$ and $P_\mathrm{M}$ are score distributions for benign and malignant masses, respectively.
    The center line for each distribution indicates the median score.
    Note that the median score for benign masses increases with assessment feedback because the averaged assessment prediction was 3.56, which is lower than that in the ground truth 3.65, leading to the overall enhancement in the score with ground truth feedback (see also \tabref{tab:2}).
    }
\end{figure}

\section{Results}
\label{sec:misc}
We first confirmed that multi-task learning improved the stand-alone accuracy of malignant lesion detection compared to single-task learning without auxiliary tasks (\tabref{tab:1}).
Note that the accuracy of malignancy detection was not increased in some multi-task models, e.g., with shape classification as an auxiliary task for the CBIS-DDSM.
This behavior may be because the mass shape in mammography has a relatively weak correlation with the malignancy\cite{Thomassin-Naggara2014}.
In addition, since the prediction on the auxiliary tasks including shape classification is correlated with the ground truth (\tabref{tab:1a}), the failure to improve the diagnosis accuracy is not due to the performance on the auxiliary tasks.

We next evaluated the accuracy with feedback on auxiliary tasks.
The existing method, REPL, on the joint multi-task architecture, failed to increase the accuracy in most models except for those with assessment classification (\tabref{tab:1}a, bottom).
This observation implies that REPL does not work well without the auxiliary task directly associated with diagnosis, such as assessment classification.
In contrast, MGA successfully improved both simple and joint multi-task models with various auxiliary tasks, including margin and distortion classifications (\tabref{tab:1}a).
The effect by MGA was also observed in malignancy score distributions for benign and malignant masses (\figref{fig:2}), where MGA pulled apart the both distribution and enhanced diagnostic performance.
Moreover, combining of MGA and REPL further increased the model's accuracy with assessment classification (\tabref{tab:1}a, bottom).
We also confirmed that MGA performs for highly suspicious nodule detection in the lung CT dataset as well (\tabref{tab:1}b).
\begin{table}[b]
\caption{\label{tab:2} 
The averaged shift on the malignancy score $\bm{s}_\mathrm{m}$ with each combination of the original prediction $\bm{y}_i$ and the feedback $\tilde{\bm{y}_i}$ for the CBIS-DDSM (mammography).
Blue and red fill indicate an increase and decrease in the score, respectively
We analysed the simple multi-task model with multiple auxiliary tasks, including margin, distortion, and assessment classifications (the model named as mix(M+D+A) in the upper table in \tabref{tab:1}).
The numerals in brackets represent the standard error, while `(-)' means that the number of corresponding samples in the cell is one.
Note that MGA can perform even when $\bm{y}_i$ and $\tilde{\bm{y}_i}$ are in the same class since $\mathcal{L}^\mathrm{T}_{\mathrm{a}_i}$ is calculable, corresponding to the situation where the user approves the model's prediction.
}
\begin{center}
\begin{tabular}{c} 
\includegraphics[width=0.8\linewidth]{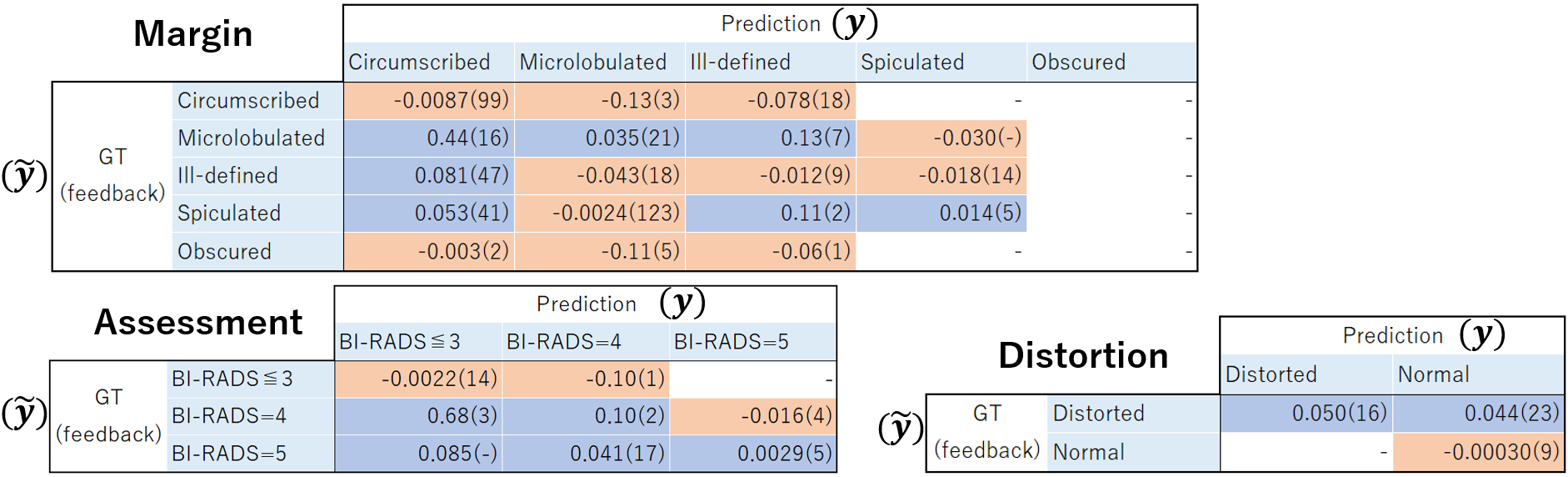}
\end{tabular}
\end{center}
\end{table}

\begin{figure}[t]
    \begin{center}
    \begin{tabular}{c} %% tabular useful for creating an array of images 
    \includegraphics[width=\linewidth]{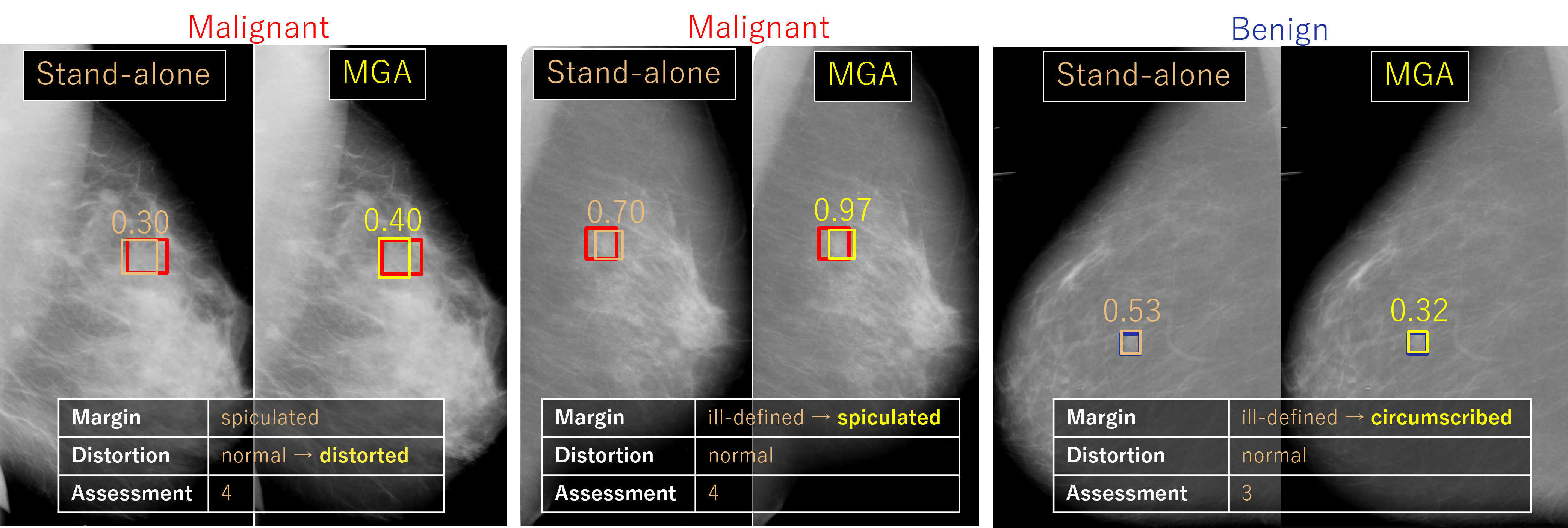}
    \end{tabular}
    \end{center}
    \caption{\label{fig:3} 
    The example cases where the feedback successfully increased and decreased the malignancy score for malignant and benign cases, respectively.
    We analyzed the simple multi-task model with multiple auxiliary tasks for CBIS-DDSM.
    The red and blue bounding boxes represent the ground truth of malignant and benign mass, respectively.
    In the bottom windows, the orange characters represent the prediction by the stand-alone multi-task model, while the yellow characters are the feedback based on the ground truth.
    See also \tabref{tab:2}.
    }
\end{figure}

\begin{figure}[t]
    \begin{center}
    \begin{tabular}{c} %% tabular useful for creating an array of images 
    \includegraphics[width=0.65\linewidth]{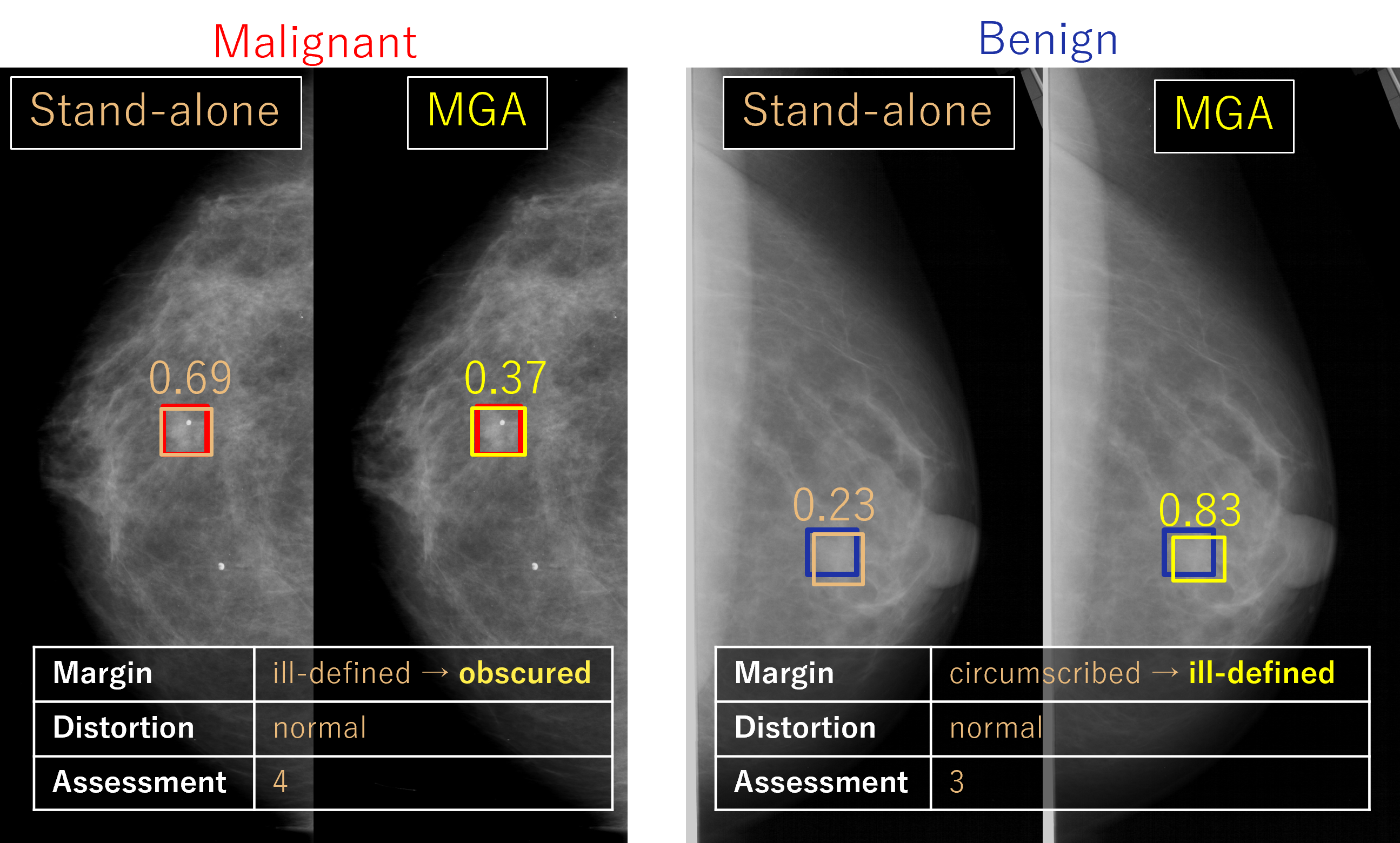}
    \end{tabular}
    \end{center}
    \caption{\label{fig:4} 
    The example cases where the feedback failed to improve the diagnostic prediction.
    We employed the same model and visualization rule as in \figref{fig:3}.
    See also \tabref{tab:2}.
    }
\end{figure}

The following analyses further underpinned the validity of MGA.
We first performed an ablation study by giving random feedback, where $\tilde{\bm{y}_i}$ was defined as a random class, to the trained multi-task model.
For models whose accuracy was improved by MGA, we found that random feedback did not perform well (\tabref{tab:1}a, top), suggesting that accurate feedback is required for adequately refining the malignancy score $\bm{s}_\mathrm{m}$.
This implication was also inspected by analyzing the averaged shift of the malignancy score $\bm{s}_\mathrm{m}$ upon feedback (\tabref{tab:2}).
We observed that the malignancy score tends to increase with specific feedback $\tilde{\bm{y}_i}$, such as a spiculated margin class and higher assessment scores, which are apparent for malignant cancers, and vice versa.
Therefore, MGA successfully reflects the characteristics of feedback on the final prediction $\bm{s}_\mathrm{m}$.

Finally, we must note that over-reliance on feedback does not always lead to improvement in the final prediction.
As shown in (\tabref{tab:2}), feedbacks with malignant and benign associated findings are likely to increase and decrease the final malignancy score, respectively.
This tendency resulted in reasonable and successful behavior in most cases as in \figref{fig:3} (see also \tabref{tab:2}). 
However, MGA sometimes worsened the diagnostic prediction, e.g., the malignancy score was increased for a benign mass when the malignancy suggestive finding was given as feedback (\figref{fig:4} right).
These behaviors imply that the MGA may over-rely on human feedback, which should be taken into account fairly and appropriately with the CAD's opinion to get better decisions. 
Further consideration is required to adaptively deal with feedback to increase the robustness of human-CAD interaction.

\section{Conclusions}
In conclusion, we proposed MGA, which allows radiologists to give feedback to the CAD system and ask for additional suggestions for diagnosis.
Our method improved the accuracy of the multi-task CAD model with the radiologist's feedback but without additional training. 
The experiments suggest that MGA can, in principle, be applied to various multi-task neural networks with branching architectures.
The interactive CAD system thereby paves the promising way for the integration of artificial intelligence into the image diagnosis process.

MGA modifies the encoded image feature instead of model weights to reflect the user's intervention.
Our method does not requires additional training of model using feedback from users, e.g., reinforcement learning\cite{Shakya2023}, which is not allowed for products in clinical use\cite{Brady2024}.
On the other hand, our approach might raise concerns that feedback could disconnect the feature from the image context.
To selectively enhance the feature relevant to feedback, MGA utilizes the gradient of the loss function between the feedback and the prediction with respect to the encoded feature.
Since the gradient adaptively finds relevant parts of the encoded feature, MGA can prevent, under reasonable hyperparameter settings, the contamination of the context unrelated to the image.
In other words, MGA improves the encoded feature in a similar manner to attention\cite{Santana2022}.

While we employed backpropagation calculation to update the image feature, forward calculation with a specifically designed network can input the user's feedback as well.
Recent studies in computer vision have explored suitable network architectures to receive the user's intention as a prompt and modify the final prediction\cite{Kirillov2023, Zhu2023, Luo2024}.
Althogh these prompting methods may suffer from the expensive cost of training the network with rich architectures, the interactive diagnosis system for the medical domain might be realized in a similar way.
Furthermore, as MGA was combinable with the simple forward calculation method like REPL, a prompting method can be further enhanced along with MGA.

The potential limitations of MGA are as follows.
For instance, MGA cannot be applied when the prediction on the auxiliary task is perfectly consistent with the radiologist's expectation.
Feedback on auxiliary tasks irrelevant to the main task, such as the shape classification for the breast mass, would also not work (see also \tabref{tab:1}).
As already discussed with \figref{fig:4}, MGA sometimes over-relies on the human feedback and fails to improve the decision.
Besides, we should further investigate the robustness against reader variability.
Our random ablation study suggests that the radiologist's feedback choice obviously impacts diagnostic accuracy.
By applying MGA to actual reading operations, we need quantitative exploration on, e.g., dependence on radiologist's expertise and robustness on uncommon findings that are difficult to interpret uniquely.
Moreover, clinical relevance of the performance improvement itself, i.e., whether the improvement of the CAD's accuracy influences on acutual reading operations, should be validated.

Finally, further consideration is required regarding which scenes the interactive diagnosis system can be adapted.
For example, our system may be more suitable for pre-treatment examinations, where precise decisions are desired, than screening where the immediate decision is preffered.
Therefore, it would be meaningful to validate our method in situations where the main task is, e.g., prognosis prediction.
In addition, MGA might contribute to multimodal disciplines such as radiogenomics\cite{Li2022}. 
Since some phenotypes are predictable from image features, we can regard genomic information prediction as an auxiliary task and utilize genetic testing data as feedback.
This framework possibly allow precise prognosis prediction and treatment guidance in the future.

\acknowledgments % equivalent to \section*{ACKNOWLEDGMENTS}       
We thank K. Watanabe, S. Namiki, R. Shino, and M. Saiko for fruitful discussions.  

% References
\bibliography{main} % bibliography data in report.bib
\bibliographystyle{spiebib} % makes bibtex use spiebib.bst

\end{document}